\documentclass[lettersize,journal]{IEEEtran}
\usepackage{amsmath,amsfonts}
\usepackage{algorithmic}
\usepackage{array}
\usepackage[caption=false,font=normalsize,labelfont=sf,textfont=sf]{subfig}
\usepackage{textcomp}
\usepackage{stfloats}
\usepackage{url}
\usepackage{verbatim}
\usepackage{graphicx}
\def\BibTeX{{\rm B\kern-.05em{\sc i\kern-.025em b}\kern-.08em
    T\kern-.1667em\lower.7ex\hbox{E}\kern-.125emX}}
\usepackage{balance}
\usepackage{amssymb}
\usepackage{subcaption}
\usepackage{tikz}
\usepackage{tikzscale}
\usepackage{pgfplots}
\usepackage{booktabs}
\usepackage{multirow}
\usepackage{hyperref}
\usepackage{threeparttable}
\usetikzlibrary{external}
\usetikzlibrary{graphs,calc,fit,backgrounds,positioning,intersections,fillbetween}

\usepackage{calc}

\usepackage{amsthm}

\usepackage{cite}

\makeatletter
\let\NAT@parse\undefined
\makeatother
\usepackage{hyperref}

\newcommand{\yihan}[1]{{#1}}
\newcommand{\revise}[1]{{#1}}

\title{
Learning to Drift \yihan{with Individual Wheel Drive}: \\ Maneuvering Autonomous Vehicle at the Handling Limits
}

\begin{document}

\author{Yihan Zhou$^{1}$, Yiwen Lu$^{1}$, Bo Yang$^{1}$, Jiayun Li$^{1}$, and Yilin Mo$^{1}$
\thanks{This work was supported by the National Natural Science Foundation of China under Grants 62461160313, 62273196, 62192752 and the BNRist project (No. BNR2024TD03003). \textit{(Corresponding Author: Yilin Mo.)}}

\thanks{$^{1}$Y.Zhou, Y.Lu, B.Yang, J.Li and Y.Mo are with the Department of Automation and BNRist, Tsinghua University, Beijing, China {\tt\footnotesize\{zhouyh23, luyw20, yang-b21, lijiayun22\}@mails.tsinghua.edu.cn, ylmo@tsinghua.edu.cn}.
}
}

\maketitle

\begin{abstract}
Drifting, characterized by controlled vehicle motion at high sideslip angles, is crucial for safely handling emergency scenarios at the friction limits.
While recent reinforcement learning approaches show promise for drifting control, they struggle with the significant simulation-to-reality gap, as policies that perform well in simulation often fail when transferred to physical systems.
In this paper, we present a reinforcement learning framework with GPU-accelerated parallel simulation and systematic domain randomization that effectively bridges the gap.
The proposed approach is validated on both simulation and a custom-designed and open-sourced 1/10 scale \yihan{Individual Wheel Drive (IWD)} RC car platform featuring independent wheel speed control.
Experiments across various scenarios from steady-state circular drifting to direction transitions and variable-curvature path following demonstrate that our approach achieves precise trajectory tracking while maintaining controlled sideslip angles throughout complex maneuvers in both simulated and real-world environments.
\end{abstract}



\section{Introduction}

In the realm of motorsport, high-speed cornering with significant sideslip angles, commonly referred to as drifting, represents an attractive yet challenging skill mastered by professional drivers \cite{cai2020high}.
It demands not only a profound understanding of vehicle dynamics but also the agility to respond instantaneously to changing environmental conditions.
Beyond serving as a thrilling spectacle, drifting constitutes a fundamental technique for trajectory tracking at the limits of vehicle handling \cite{betz2022autonomous}. As autonomous vehicles are expected to handle extreme scenarios with capabilities matching or exceeding those of human drivers, research on drifting control has substantial implications for the safety of autonomous vehicles under critical conditions such as sudden changes in road friction or emergency obstacle avoidance \cite{goh2020toward,yang2022hierarchical,lu2023consecutive}.

\begin{figure}[h!]
    \centering
    \includegraphics[width = \columnwidth]{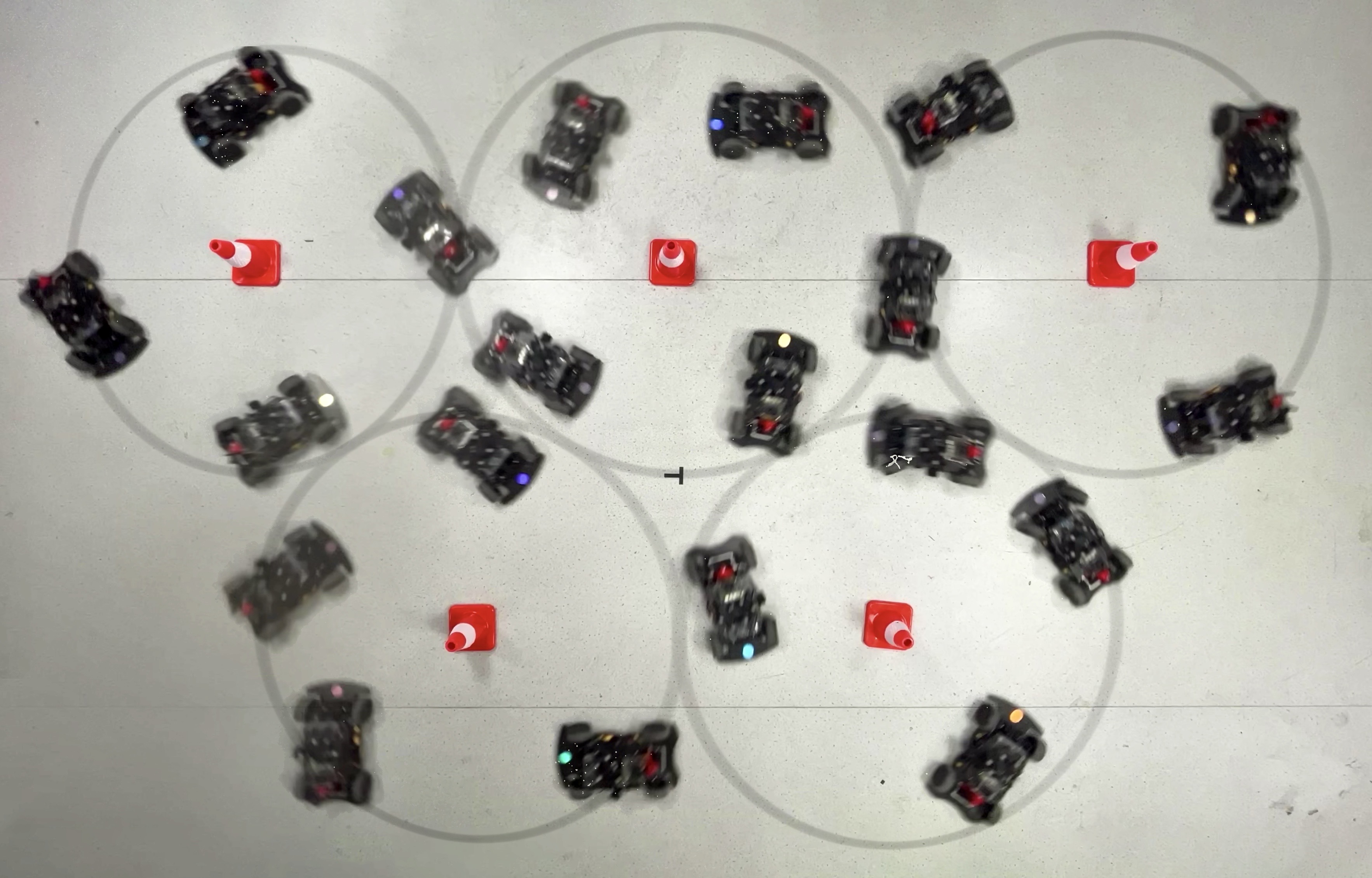}
    \caption{Top-down view of drifting along the path inspired by the Olympic rings. Five tangent circles form the reference path marked by gray curves, with red cones marking the centers. Multiple car positions captured in this long-exposure photograph demonstrate consecutive drifting maneuvers.}
    \label{fig:top-down}
    \vspace{-0.5cm}
\end{figure}

Several works have explored autonomous drifting through various approaches. 
Model-based methods leverage vehicle dynamics principles \cite{voser2010analysis,hindiyeh2014controller,goh2020toward}, typically identifying vehicle parameters and then formulating drifting as optimal control problems \cite{weber2023modeling} or designing modular control architectures \cite{goh2020toward}.
While these approaches provide interpretability and theoretical guarantees, they face practical limitations including computational complexity and sensitivity to model uncertainties. 

Learning-based methods have emerged as a complementary paradigm. Recent work by Djeumou et al.~\cite{djeumou2024one} demonstrated how learned dynamics can enhance control by incorporating a diffusion model into an MPC framework, achieving robust drifting across varying vehicles and conditions.
In parallel, reinforcement learning (RL) represents another promising branch, with early work by Cutler et al.~\cite{cutler_autonomous_2016} demonstrating feasibility for circular drift stabilization on small-scale platforms, though with limited trajectory flexibility. Subsequent research by Cai et al.~\cite{cai2020high} and Domberg et al.~\cite{domberg_deep_2022} extended RL capabilities to high-speed racing scenarios in simulation.
However, these approaches encounter significant limitations that hinder practical deployment. The simulation-to-reality gap remains particularly pronounced in drifting control, as evidenced in \cite{domberg_deep_2022} where policies performing well in simulation failed to complete maneuvers when transferred to physical platforms. Additionally, conventional RL implementations often require extensive training time, with approaches like \cite{cai2020high} requiring over 11 hours of training time, extending the development and tuning cycle of the algorithm.
Consequently, there exists a need for a framework that can both accelerate the training process and effectively bridge the sim-to-real gap, enabling robust drifting control across diverse trajectory patterns.

Research platforms for drifting control have predominantly utilized either rear-wheel drive (RWD) \cite{goh2016simultaneous, meijer2024nonlinear} or all-wheel drive (AWD) \cite{yang2022hierarchical, lu2023consecutive, 10219145} with coupled wheel speeds.
The advancement of electric vehicles has accelerated the development of Individual Wheel Drive (IWD) architectures that enable independent control of each wheel's speed and torque, \yihan{expanding the available control authority compared to conventional drivetrains.} However, despite their increasing prevalence in the automotive industry, research platforms for developing and validating IWD control strategies remain limited.

In this paper, we present an integrated hardware-software solution that addresses the aforementioned limitations in autonomous drifting control.
From an algorithmic perspective, we propose a reinforcement learning framework that successfully bridges the simulation-to-reality gap through systematic domain randomization techniques. \yihan{This framework leverages custom GPU-accelerated parallel simulation, reducing training time from hours to minutes.}
From the hardware perspective, our methodology is implemented on a custom-designed Individual Wheel Drive (IWD) platform that we have developed and open-sourced, delivering enhanced control authority through independent wheel actuation.
The integration enables the exploration of complex drift maneuvers while maintaining computational efficiency suitable for embedded systems. 
Through experimental validation, we demonstrate that the approach achieves precise drifting control across diverse trajectories, including challenging scenarios such as eight-shaped pattern and variable-curvature track, with consistent performance in both simulation and real-world environments.

\emph{Contributions:} This work advances autonomous drifting control through:
\yihan{(1) A reinforcement learning approach for autonomous drifting control on IWD vehicles that enables effective sim-to-real transfer without real-world fine-tuning;}
(2) An open-source IWD platform implementation providing enhanced maneuverability through independent wheel speed control with precise torque vectoring capabilities;
and \yihan{(3) Experimental demonstration of challenging autonomous drifting maneuvers including direction reversals and variable-curvature tracking.}
The complete implementation is publicly available to promote reproducibility and further research in this domain.

The remainder of this paper is organized as follows:
Section~\ref{sec:related} reviews the related work in autonomous drifting.
Section~\ref{sec:model} presents the platform design and system modeling, including the IWD RC car platform, vehicle dynamics model, and the GPU-accelerated parallel car simulator.
Section~\ref{sec:method} formulates the consecutive drifting problem and details our method for agile maneuvering. Section~\ref{sec:result} presents experimental results, and Section~\ref{sec:conclusion} concludes the paper.

\section{Related Works}
\label{sec:related}

\subsection{Model-based Approaches for Drifting Control}
Autonomous drifting has been extensively studied using model-based approaches. Early works by Voser et al.~\cite{voser2010analysis} and Hindiyeh and Gerdes~\cite{hindiyeh2014controller} established fundamental frameworks by analyzing equilibrium states and developing nested-loop control structures for drift stabilization. Goh et al.~\cite{goh2016simultaneous,goh2018controller,goh2020toward} extended these concepts to arbitrary path tracking by incorporating curvilinear coordinates and nonlinear model inversion. Various optimal control techniques have been applied to drifting, including MPC \cite{hu2024novel}, NMPC \cite{shi2023nonlinear,weber2023modeling, meijer2024nonlinear}, and LQR \cite{velenis2009steady, velenis2011steady}. While these approaches offer theoretical guarantees, they face practical challenges due to their computational complexity and reliance on accurate system identification and tire modeling, though some limitations can be addressed through expert knowledge-based methods \cite{joa2020new}.

\subsection{Learning-based Methods for Drifting Control}
Recent work on learning-based approaches for autonomous drifting has shown significant progress through two main directions. Djeumou et al. enhanced model-based control frameworks by integrating the learned tire \cite{djeumou2023autonomous} or vehicle model \cite{djeumou2024one}, improving robustness to varying conditions.
Meanwhile, reinforcement learning represents an effective alternative to model-based control approaches.
Early explorations using model-based RL frameworks like PILCO \cite{deisenroth2011pilco} demonstrated initial feasibility of learning-based methods for drift control in both simulation \cite{bhattacharjee2018autonomous} and on small-scale RC cars \cite{cutler_autonomous_2016}, though their applications were limited to circular trajectories. Recent advances in model-free RL have tackled more complex scenarios: Cai et al.~\cite{cai2020high} developed a SAC-based approach combining professional driver demonstrations for high-speed drifting across various race tracks in the CARLA simulator, while Domberg et al.~\cite{domberg_deep_2022} proposed a PPO-based method for arbitrary trajectory tracking that maps vehicle states and path information directly to control commands.
However, these methods face two main challenges.
First, the simulation-to-reality gap limits the real-world performance, particularly on complex trajectories, leading to degraded performance when transferring from simulation to physical systems \cite{domberg_deep_2022}. 
Second, the training efficiency remains a practical concern, with training times exceeding 11 hours in \cite{cai2020high}, which slows the iterative refinement process.
\section{Platform and System Modeling}
\label{sec:model}
To facilitate research in agile vehicle maneuvering, we introduce an open-source 1/10 scale RC car testbed featuring independent wheel drive (IWD) capabilities. This section details our platform design and develops its mathematical model which is then leveraged in a GPU-accelerated parallel simulation environment to enable efficient reinforcement learning.

\subsection{Individual Wheel Drive RC car}
\label{sec:hardware}
\begin{figure}[!t]
    \centering
    \begin{minipage}[c]{0.45\columnwidth}
        \centering
        \includegraphics[width=0.9\linewidth]{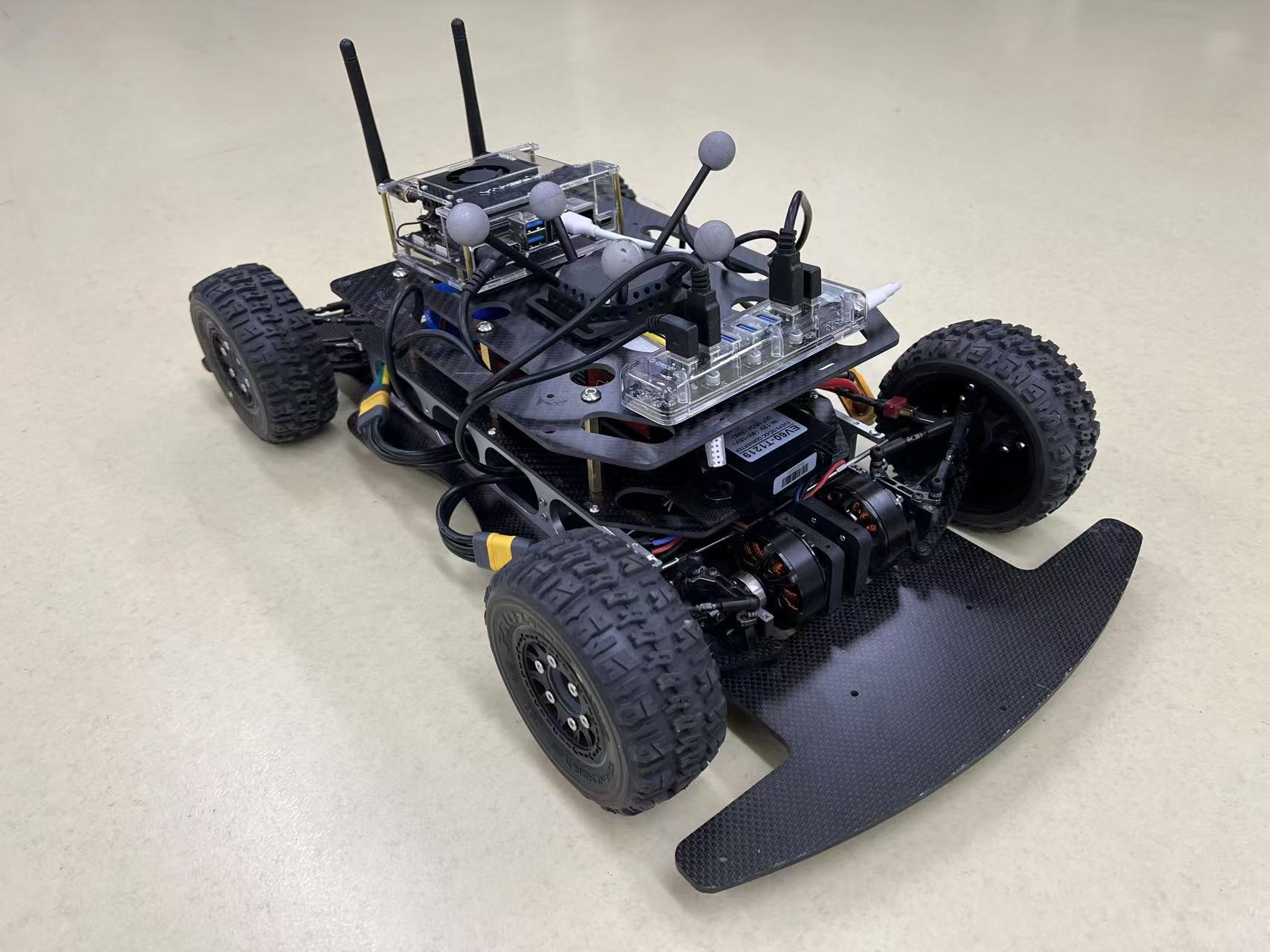}
    \end{minipage}%
    \hfill
    \yihan{
    \begin{minipage}[c]{0.5\columnwidth}
        \centering
        \small
        \begin{tabular}{ll}
        \toprule
        Parameter & Value \\
        \midrule
        Vehicle mass & 4.84 kg \\
        Wheelbase & 0.35 m \\
        Track width & 0.26 m \\
        Wheel radius & 0.0565 m \\
        Max steering angle & $\pm$26.35$^{\circ}$ \\
        \bottomrule
        \end{tabular}
    \end{minipage}
    }
    \caption{Xcar platform and key specifications.}
    \label{fig:xcar_platform}
    \vspace{-0.3cm}
\end{figure}

Our experimental platform, Xcar (shown in Fig.~\ref{fig:xcar_platform}), draws inspiration from several established research vehicles including the Berkeley autonomous race car \cite{gonzales2018planning}, MIT RACECAR \cite{karaman2017project}, MuSHR \cite{srinivasa2019}, RoSCAR \cite{Hart2014} and F1TENTH \cite{okelly2019}.
Unlike previous platforms, Xcar incorporates a distinct drivetrain configuration with four independent motors powered by a 4S LiPo battery, each controlled by a VESC \yihan{Mini V6.7} motor controller, which provides additional degrees of freedom for vehicle dynamics control enabling precise torque vectoring and enhanced yaw moment control.
The onboard computational system utilizes an NVIDIA Jetson Orin Nano, serving as the ROS master node and executing the control policy.
To establish accurate localization and state estimation, the car is equipped with a motion capture system.
Using this high-precision pose data, a Kalman filter is implemented to estimate linear and angular velocities and accelerations.
A detailed list of hardware components is available at \url{https://zhou-yh19.github.io/xcar-hardware}.

\subsection{Modeling}
\label{sec:vehicle_model}
To balance model fidelity with computational efficiency, we adopt a simplified vehicle dynamics model based on 7-DOF models \cite{milani2021vehicle,velenis2009steady} by constraining motion to a 2D plane and omitting motor dynamics, while retaining the essential Pacejka tire model.

\begin{figure}[!t]
    \centering
    \includegraphics[width=0.28\textwidth]{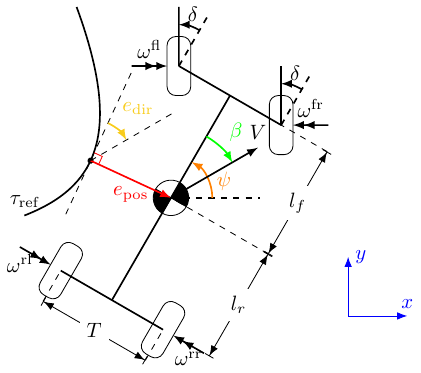}
    \caption{Illustration of IWD vehicle model, the reference trajectory $\tau_{\mathrm{ref}}$ and the definition of errors.}
    \label{fig:model_xcar}
\end{figure}

The IWD vehicle's state vector $\mathbf{x} \in \mathcal{X}$ consists of six components representing its position and motion (illustrated in Fig. \ref{fig:model_xcar}):
\begin{equation}
\mathbf{x} = [x, y, \psi, \dot{x}, \dot{y}, \dot{\psi}],
\end{equation}
where $(x, y)$ denotes the vehicle's position in global coordinates, $\psi$ represents the heading angle, and $(\dot{x}, \dot{y}, \dot{\psi})$ are their respective time derivatives.

The control input vector $\mathbf{u} \in \mathcal{U}$ encompasses the steering angle and individual wheel angular velocities:
\begin{equation}
\mathbf{u} = [\delta, \omega^\text{fl}, \omega^\text{fr}, \omega^\text{rl}, \omega^\text{rr}],
\end{equation}
where $\delta$ represents the steering angle and $\omega^\text{ij}$ denotes the angular velocity of each wheel (subscripts: f/r - front/rear, l/r - left/right).

\revise{
The dynamics of the IWD vehicle are governed by the following equations:
\begin{equation*}
\begin{aligned}
m \ddot{x} = & \cos(\delta+\psi)F_{x}^{\text{f}} - \sin(\delta+\psi)F_{y}^{\text{f}} + \cos\psi F_{x}^{\text{r}} - \sin\psi F_{y}^{\text{r}} \\
m \ddot{y} = & \sin(\delta+\psi)F_{x}^{\text{f}} + \cos(\delta+\psi)F_{y}^{\text{f}} + \sin\psi F_{x}^{\text{r}} + \cos\psi F_{y}^{\text{r}} \\
I_z \ddot{\psi} = & [\cos \delta F_{y}^{\text{f}} + \sin \delta F_{x}^{\text{f}}] l_f - F_{y}^{\text{r}} l_r \\
& + [\cos \delta \Delta F_{x}^{\text{f}} - \sin \delta \Delta F_{y}^{\text{f}} + \Delta F_{x}^{\text{r}}] T/2
\end{aligned}
\label{eq:dynamics}
\end{equation*}
where $m$ is the vehicle mass, $I_z$ is the moment of inertia, $l_f$ and $l_r$ are the distances from the center of mass to the front and rear axles, $T$ is the track width. The force terms are defined as: $F_{x/y}^{\text{f}} = F_{x/y}^{\text{fl}} + F_{x/y}^{\text{fr}}$ (front axle forces), $F_{x/y}^{\text{r}} = F_{x/y}^{\text{rl}} + F_{x/y}^{\text{rr}}$ (rear axle forces), $\Delta F_{x/y}^{\text{f}} = F_{x/y}^{\text{fr}} - F_{x/y}^{\text{fl}}$ (front differential forces), and $\Delta F_{x/y}^{\text{r}} = F_{x/y}^{\text{rr}} - F_{x/y}^{\text{rl}}$ (rear differential forces), with $F_{x/y}^\text{ij}$ representing the longitudinal/lateral tire forces at each wheel.
}
\yihan{These tire forces are derived from the control inputs through the Pacejka tire model based on wheel slip and vertical loads.}
\yihan{Our implementation includes longitudinal load transfer but assumes equal lateral distribution.}

For analyzing and controlling drift maneuvers, three coordinate-free variables play a fundamental role:
\begin{itemize}
    \item The sideslip angle $\beta = \tan^{-1}\left(\dfrac{\dot{y}}{\dot{x}}\right)-\psi.$
    \item The yaw rate $r$ (equivalent to $\dot{\psi}$).
    \item The velocity magnitude $V = \sqrt{\dot{x}^2 + \dot{y}^2}.$
\end{itemize}

These variables relate to the vehicle motion through the kinematic equations:
\begin{equation}
\dot{x}=V \cos (\beta+\psi), \quad \dot{y}=V \sin (\beta+\psi), \quad \dot{\psi}=r.
\end{equation}
Therefore, $[r, \beta, V]$ can be viewed as a rotation- and translation-invariant description of the vehicle motion, and therefore particularly important for learning algorithm design.

\subsection{GPU-accelerated parallel car simulation}
\label{sec:simulator}
Based on the vehicle model described above, we develop a GPU-accelerated car simulator inspired by NVIDIA Isaac Gym \cite{makoviychuk2021isaac} to facilitate efficient reinforcement learning training. 
\yihan{The simulator supports various drivetrain configurations including IWD, RWD, and AWD.}
Compared to traditional CPU-based simulators like CarSim\textsuperscript{\small\textregistered} and F1TENTH \cite{o2020f1tenth}, our simulator offers two key advantages: (1) enabling massive parallelization on a single machine to meet the data requirements of deep reinforcement learning, and (2) eliminating CPU-GPU data transfer overhead during neural network training.

The simulator is implemented in PyTorch \cite{paszke2017automatic} using the Euler integration method. Running on an NVIDIA GeForce RTX 3080 GPU with a 0.01s time step, it achieves parallel simulation of $10^6$ car instances at $30$ steps per second, corresponding to a speedup factor of $3 \times 10^5$ relative to real-time.

This high-throughput approach prioritizes rapid data generation over absolute fidelity, acknowledging that even high-fidelity models face reality gaps due to parametric uncertainties. The task of bridging this gap is addressed through domain randomization techniques during reinforcement learning training, as detailed in Section~\ref{sec:domain_randomization}.
\section{Reinforcement Learning for Maneuvering}
\label{sec:method}

\begin{figure*}[t]
    \centering
   \includegraphics[width=0.9\textwidth]{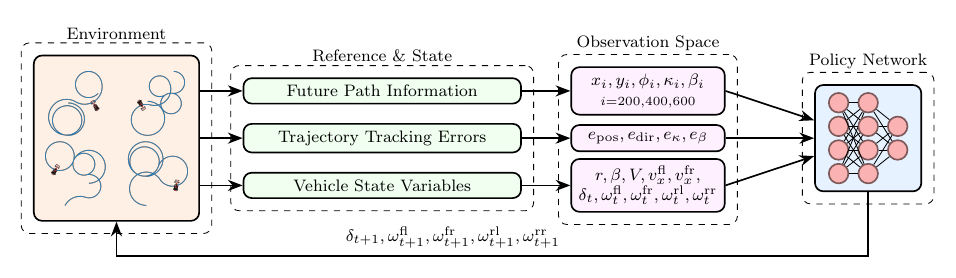}
   \caption{Overview of the reinforcement learning framework for autonomous drifting: The GPU-accelerated parallel simulation environment accelerates computation with multiple vehicle instances across various trajectories. For coordinate-free policy learning, the observation space is constructed from translation and rotation invariant quantities including vehicle velocities and slip angles, along with tracking errors and future waypoints computed in a curvilinear coordinate system aligned with the reference trajectory, leading to end-to-end control of individual wheel speeds and steering angles.}
   \label{fig:diagram}
   \vspace{-0.3cm}
\end{figure*}

In this section, we present a reinforcement learning framework (Fig.~\ref{fig:diagram}) that enables autonomous vehicles to perform agile drifting maneuvers at the handling limits while addressing the fundamental challenge of sim-to-real transfer. The framework trains a policy network that controls individual wheel speeds and steering angle for precise trajectory tracking and stable drift behavior. We design systematic domain randomization strategies across reference trajectories, initial conditions, \yihan{tire model parameters,} and dynamic disturbances to bridge the gap between simulation and physical implementation.

\subsection{Problem Formulation}

We formulate autonomous drifting control as a Markov Decision Process (MDP) $\mathcal{M}=(\mathcal{S}, \mathcal{A}, \mathcal{P}, \mathcal{R}, \gamma)$, where the state space $\mathcal{S}$ encompasses both vehicle states and reference trajectory information, $\mathcal{A}$ represents the action space including steering angle and individual wheel speeds, $\mathcal{P}$ denotes the transition dynamics governed by the IWD vehicle model, $\mathcal{R}$ is the reward function designed to achieve precise trajectory tracking, stable drift performance with significant sideslip angles, and smooth control quality, $\gamma$ is the discount factor.
The objective is to find a policy $\pi: \mathcal{S} \rightarrow \mathcal{A}$ that maps states to control actions while maximizing the expected return:

\begin{equation}
    J(\pi)= \mathbb{E}_\pi\left[\sum_{t=0}^{\infty} \gamma^t r\left(s_t, a_t\right)\right].
\end{equation}

\subsection{Learning Framework}

Building upon the problem formulation, we specify the concrete design of each component in our framework, including a carefully designed observation space, a physically constrained action space, and a structured reward function.

\subsubsection{Observation space}

While the MDP state space $\mathcal{S}$ encompasses the complete vehicle dynamics and environmental information, direct policy learning from raw states often suffers from sampling inefficiency, primarily due to dimensionality issues and lack of invariance properties. Therefore, we design an observation space $\mathcal{O}$ that extracts the essential information from $\mathcal{S}$ and transforms it into a representation efficient for policy learning, which includes:

\begin{itemize}
\item \textbf{Future Waypoints Information} enables the policy to anticipate path curvature changes and adjust control inputs for smoother transitions. For each preview point along the reference path, we express its position, heading angle, and sideslip angle in the vehicle frame.
\item \textbf{Trajectory Tracking Errors} measure the vehicle's deviation from the reference trajectory, including lateral position error, heading angle error, path curvature discrepancy, and sideslip angle deviation.
\item \textbf{Vehicle State Variables} describe the instantaneous dynamics through coordinate-independent quantities comprising yaw rate $r$, sideslip angle $\beta$, velocity $V$, wheel velocities, and previous control inputs ($\delta$ and $\omega^\text{ij}$).
\end{itemize}

These observations provide the policy with comprehensive state information, enabling it to learn effective control strategies for stable drifting along the reference trajectories.

\subsubsection{Action space}

The steering angle is bounded to $[-0.46, 0.46]$ radians (approximately $\pm 26.35^\circ$), corresponding to the physical limits of the steering mechanism. The wheel linear velocities are constrained within $[1,7]~\text{m}/\text{s}$, where the lower bound prevents the vehicle from stopping or moving too slowly to maintain drift conditions, while the upper bound ensures controllable drifting behavior.

\subsubsection{Reward Design}

The reward function integrates multiple objectives to achieve stable drift behavior while ensuring precise trajectory following:
\begin{equation}
r = r_\text{tracking} + r_\text{control} + r_\text{aux}.
\end{equation}

Trajectory tracking rewards $r_\text{tracking}$ quantify the vehicle's tracking accuracy relative to the reference path through quadratic penalties on tracking errors:
\begin{equation}
r_\text{tracking} = w_\text{pos}r_\text{pos} + w_\text{dir}r_\text{dir} + w_\text{curv}r_\text{curv} + w_\text{drift}r_\text{drift},
\end{equation}
where:
\begin{equation}
\begin{aligned}
& r_\text{pos} = -e_\text{pos}^2, r_\text{dir} = -e_\text{dir}^2, r_\text{curv} = -e_\kappa^2 =  -(\kappa - \kappa_\text{ref})^2, \\ & r_\text{drift} = -e_\beta^2 =  -(\beta - \beta_\text{ref})^2.
\end{aligned}
\end{equation}

Here, $e_\text{pos}$ and $e_\text{dir}$ represent position and direction errors defined in a curvilinear coordinate system along the reference trajectory as shown in Fig.~\ref{fig:model_xcar}, $\kappa$ and $\kappa_\text{ref}$ represent the actual and reference path curvature, and $\beta$ and $\beta_\text{ref}$ denote the actual and reference sideslip angles.

Control quality terms $r_\text{control}$ impose constraints on the control inputs to ensure vehicle stability:
\begin{equation}
r_\text{control} = w_\text{smooth}r_\text{smooth}.
\end{equation}

The smoothness term $r_\text{smooth}$ penalizes rapid changes in consecutive commands:
\begin{equation}
r_\text{smooth} = -(\delta_t - \delta_{t-1})^2 - 10^{-4} \sum_{ij \in \{\text{fl,fr,rl,rr}\}} (\omega^\text{ij}_{t} - \omega^\text{ij}_{t-1})^2,
\end{equation}
where subscripts $t$ and $t-1$ denote current and previous time steps, and the scaling factor $10^{-4}$ balances the magnitudes of wheel speed variations relative to steering changes.

To further shape the learning process and ensure proper drift execution, we incorporate auxiliary terms:
\begin{equation}
r_\text{aux} = w_\text{slip}r_\text{slip} + w_\text{speed}r_\text{speed} + w_\text{prog}r_\text{prog}.
\end{equation}

The wheel slip regulation term limits longitudinal slip at the front wheels to preserve steering effectiveness:
\begin{equation}
r_\text{slip} = - (v^\text{fl}_x - \omega^\text{fl}R)^2 - (v^\text{fr}_x - \omega^\text{fr}R)^2,
\end{equation}
where $v^\text{ij}_{x}$ represents the wheel's longitudinal velocity and $R$ is the wheel radius.

A velocity incentive encourages the vehicle to maintain sufficient speed necessary for sustained drift execution:
\begin{equation}
    r_\text{speed} = \min(0, V-0.5).
\end{equation}

In addition, to encourage stable and consistent forward movement along the reference trajectory, we incorporate a progression reward $r_{\text{prog}}$ that is proportional to the distance traveled along the reference trajectory, clipped to a maximum value corresponding to a fixed distance. This reward term incentivizes the agent to maintain forward progress while preventing excessive rewards from the policy exploiting shortcuts.

\subsection{Domain Randomization Strategy}
\label{sec:domain_randomization}
To bridge the simulation-to-reality gap in autonomous drifting, we implement domain randomization techniques that address key uncertainties in real-world vehicle dynamics.

\subsubsection{Trajectory Randomization for Policy Generalization}
We create continuous paths by integrating curvature profiles with varying signs and magnitudes, maintaining smoothness ($C^{(1)}$ continuity) across transitions, which enables the policy to learn both stable drifting and smooth transitions between opposite drift directions. Each generated path provides context through position coordinates, velocity direction, and the desired direction of sideslip angle determined by local path curvature.

\subsubsection{Initial State Randomization for Robust Learning}
To handle real-world uncertainties and unexpected states during drift execution, we implement state randomization in our training, initializing vehicles at varied positions along reference trajectories, with controlled Gaussian perturbations added to starting positions ($\Delta x, \Delta y \sim \mathcal{N}(0, \sigma_{\text{pos}}^2)$) and heading angles ($\Delta \psi \sim \mathcal{N}(0, \sigma_{\text{heading}}^2)$). We further randomize dynamic states including velocities, sideslip angles, and yaw rates based on local trajectory curvature, which ensures the policy learns robust control strategies that generalize beyond training conditions.

\subsubsection{Tire Model and Dynamic Disturbance Randomization}
\yihan{To account for unmodeled dynamics and varying tire-road interactions, we implement disturbances at two levels. We randomize Pacejka tire model parameters ($B$, $C$, $D$) sampled uniformly at each episode initialization to represent different tire-road conditions. We further implement an auto-regressive disturbance process that injects structured noise into the tire forces. The disturbance follows $d_{t+1} = ad_t + w\varepsilon_t$, where $d_t$ represents the disturbance vector affecting tire forces, $a$ controls temporal correlation, and $w$ scales the Gaussian innovation $\varepsilon_t$. This dual-layer approach accounts for both persistent and transient uncertainties in tire-road interactions.}

\revise{The contribution of each randomization component is analyzed through ablation studies (see Appendix).}

\section{Experimental Results}
\label{sec:result}

We validate our approach through extensive experiments in both simulation and real-world environments, demonstrating the policy's ability to handle different drifting scenarios. The code is available at \url{https://github.com/zhou-yh19/xcar-rlgpu}.

\subsection{Training Setup}
\yihan{We train the policy using Proximal Policy Optimization (PPO) algorithm \cite{schulman2017proximal}, known for its training stability and requiring little hyperparameter tuning \cite{schulman2015trust}.}
The policy network consists of three fully connected layers with (64, 32, 16) neurons respectively. The reference trajectories are parameterized as a sequence of waypoints with a fixed distance of $0.005\text{m}$ between consecutive points.
Our implementation utilizes approximately \( 10^5 \) parallel simulation environments on an NVIDIA GeForce RTX 3080 GPU, achieving convergence within 10.762 minutes on average.
The training parameters are summarized in Table~\ref{tab:training_params}.

\begin{table}[!t]
    \renewcommand{\arraystretch}{1.2}
    \caption{Training Parameters}
    \label{tab:training_params}
    \centering
    \begin{tabular}{p{0.55cm}c|p{0.55cm}c|p{0.5cm}c|p{0.4cm}cp{1.2cm}c}
    \toprule
    \multicolumn{4}{c|}{\textbf{Reward Weights}} & \multicolumn{4}{c}{\textbf{Randomization Parameters}} \\
    \midrule
    Reward & Weight & Reward & Weight & Param. & Value & Param. & Range \\
    \midrule
    $r_\text{pos}$ & 2.4 & $r_\text{speed}$ & 0.1 & $\sigma_\text{pos}$ & 0.1\,m & $r_\text{init}$ & [1,3]\,rad/s\\
    $r_\text{dir}$ & 0.5 & $r_\text{smooth}$ & 0.015 & $\sigma_\text{heading}$ & 0.1\,rad & $\beta_\text{init}$ & [-1,1]\,rad \\
    $r_\text{curv}$ & 0.15 & $r_\text{slip}$ & 0.005 & $a$ & 0.95 & $V_\text{init}$ & [0,3]\,m/s \\
    $r_\text{drift}$ & 1.6 & $r_\text{prog}$ & 0.2 & $w$ & 0.1 & &\\
    \bottomrule
    \end{tabular}
    \begin{tablenotes}
        \footnotesize
        \item \yihan{Tire model parameters: $B \in [0.8, 1.0]$, $C \in [2.0, 2.5]$, $D \in [0.3, 0.4]$}
    \end{tablenotes}
\end{table}

\subsection{Experimental Validation}

We evaluate the trained policy across three distinct trajectory patterns:
\begin{itemize}
    \item \textbf{Steady-State Circular Drift:} Basic drift stabilization along a circular path (radius = 1\,m)
    \item \textbf{Eight-Shaped Maneuver:} Controlled drift through trajectory intersections requiring drift direction reversals
    \item \textbf{Variable-Curvature Track:} Advanced trajectory featuring varying path curvature while maintaining consistent drift states
\end{itemize}

\subsection{Simulation Results}

The policy demonstrates robust performance across all trajectory patterns while maintaining significant sideslip angles (45°-55°) and stable velocities (1.5-2.5\,m/s). Detailed analysis of each trajectory type reveals the following characteristics:

\begin{figure}[!t]
    \centering
    \begin{minipage}[c]{0.48\linewidth}
    \centering
    \includegraphics[width=0.9\columnwidth]{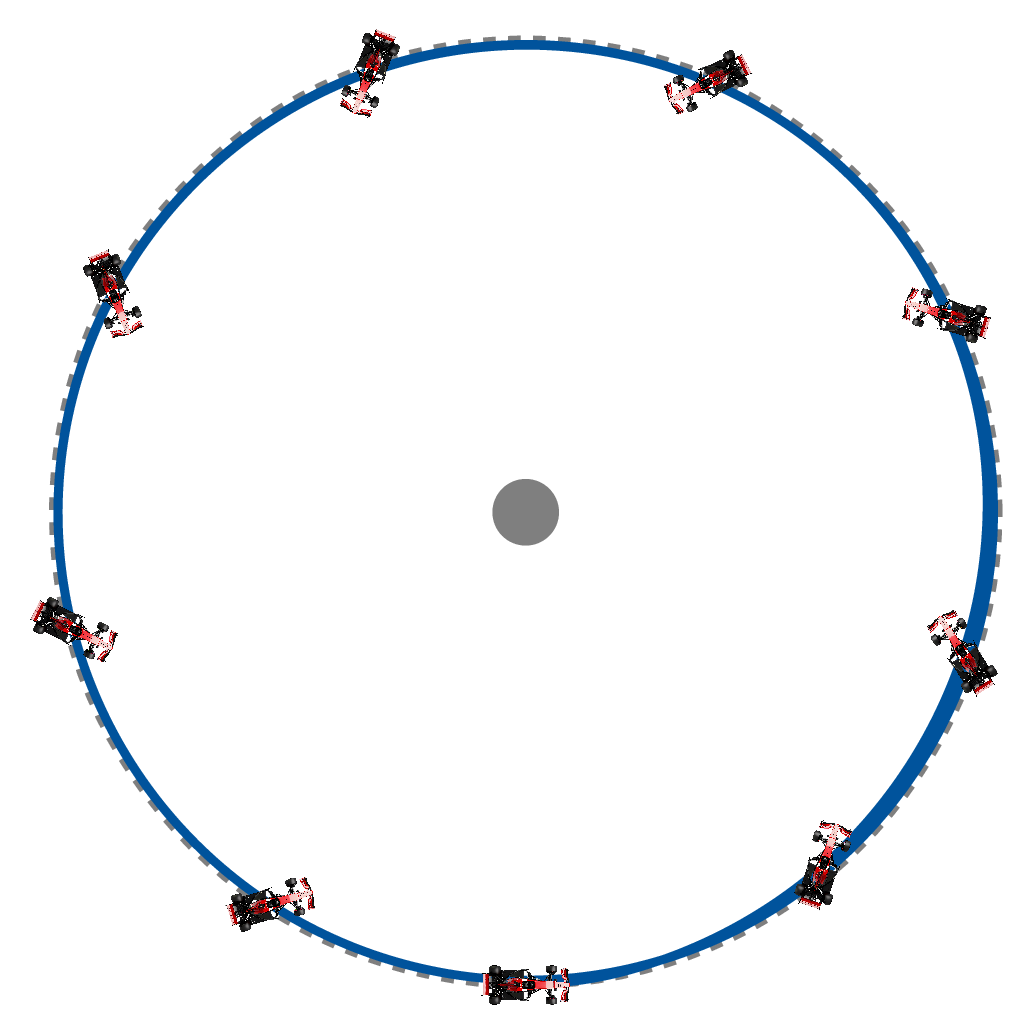}
    \caption*{(a)}
    \vspace{0.2cm}
    \includegraphics[width=\columnwidth]{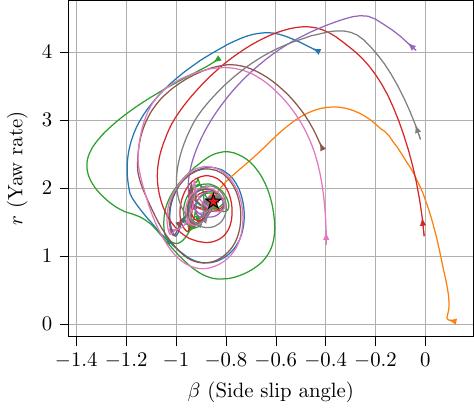}
    \caption*{(b)}
    \end{minipage}%
    \hfill
    \begin{minipage}[c]{0.48\linewidth}
    \centering
    \includegraphics[width=\columnwidth]{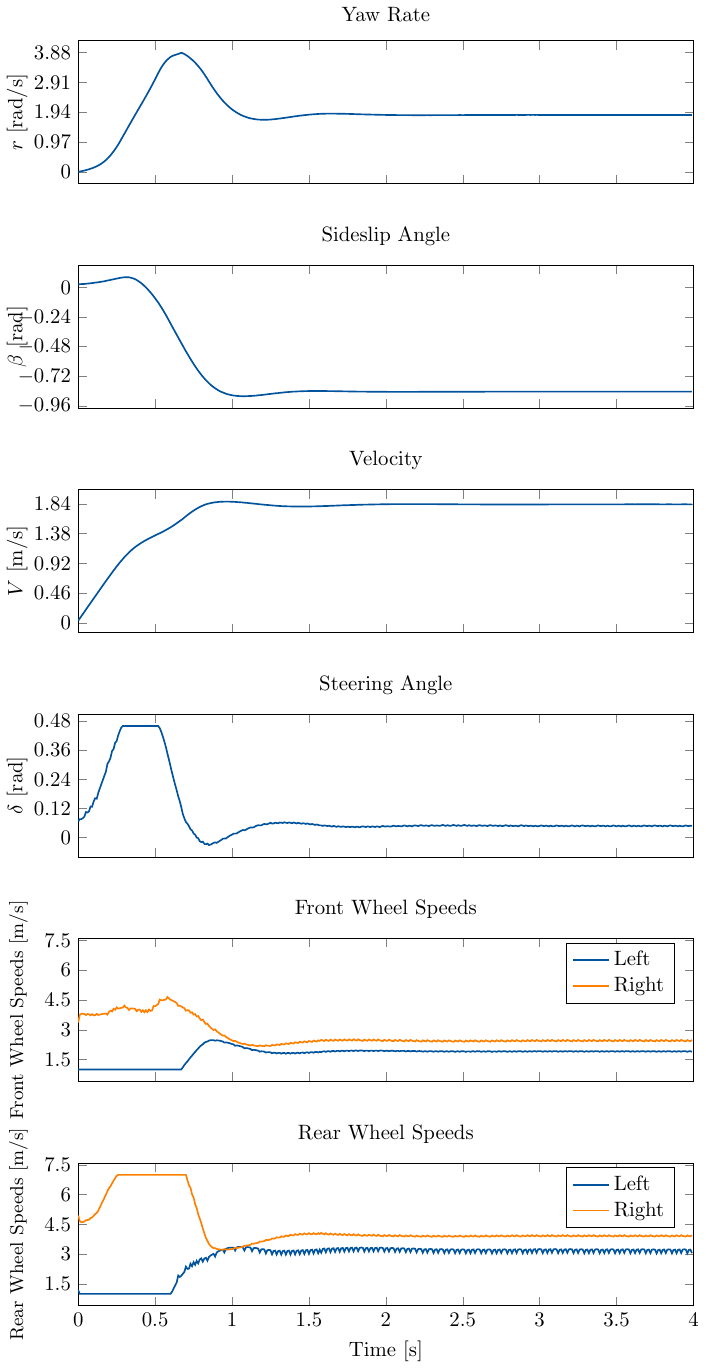}
    \caption*{(c)}
    \end{minipage}
    \vspace{-0.2cm}
    \caption{Performance of the IWD drifting policy on a circular trajectory. (a) Paths for circular trajectory tracking (blue for actual vehicle path, gray for reference circle); filled gray circle marks the center, race cars show vehicle poses at sampled positions. (b) State convergence in the $r$-$\beta$ phase plane, showing trajectories from different initial conditions converging to a natural drift equilibrium. (c) Time histories of states ($r$, $\beta$, $V$) and control inputs from zero initial conditions.}
    \label{fig:drift}
    \vspace{-0.4cm}
\end{figure}

Fig.~\ref{fig:drift} shows the performance for circular trajectory (radius = 1\,m). 
The controller achieves precise tracking with position errors below 0.1~m, as observed in Fig.~\ref{fig:drift}(a).
Fig.~\ref{fig:drift}(b) illustrates the convergence behavior in the $r$-$\beta$ plane.
Despite starting from various initial conditions, the system consistently converges to a natural drift equilibrium ($r = 1.85$\,rad/s, $\beta = -0.85$\,rad, $V = 1.84$\,m/s), which suggests that the learned policy has discovered a stable operating point that satisfies both the vehicle dynamics constraints and the geometric requirements for maintaining circular drifting.

We further analyze the time histories of states and control inputs from zero initial conditions in Fig.~\ref{fig:drift}(c), which illustrates the distinct characteristics of IWD control during drift maneuvers.
During the initiation phase (0--1.0\,s), the controller exploits differential wheel speeds to induce drift efficiently - the rear right wheel maintains significantly higher angular velocity compared to the rear left wheel, with a similar but less pronounced differential on the front wheels.
Once the system reaches steady-state drifting, the controller maintains asymmetric wheel speeds across the left and right sides of the vehicle, but with a reduced differential magnitude.

\begin{figure}[!t]
    \centering
    \begin{minipage}[c]{0.48\linewidth}
    \centering
    \includegraphics[width=0.9\columnwidth]{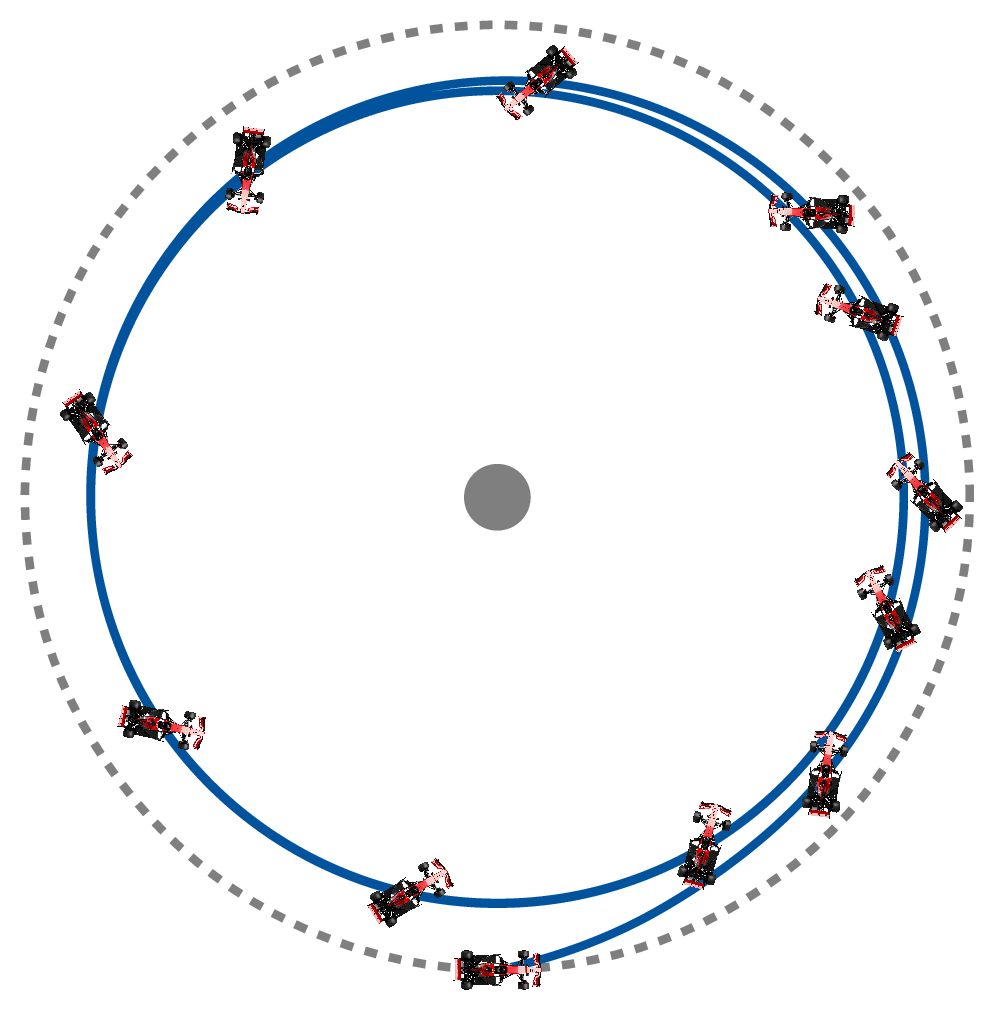}
    \caption*{(a)}
    \vspace{0.2cm}
    \end{minipage}%
    \hfill
    \begin{minipage}[c]{0.50\linewidth}
    \centering
    \includegraphics[width=0.9\columnwidth]{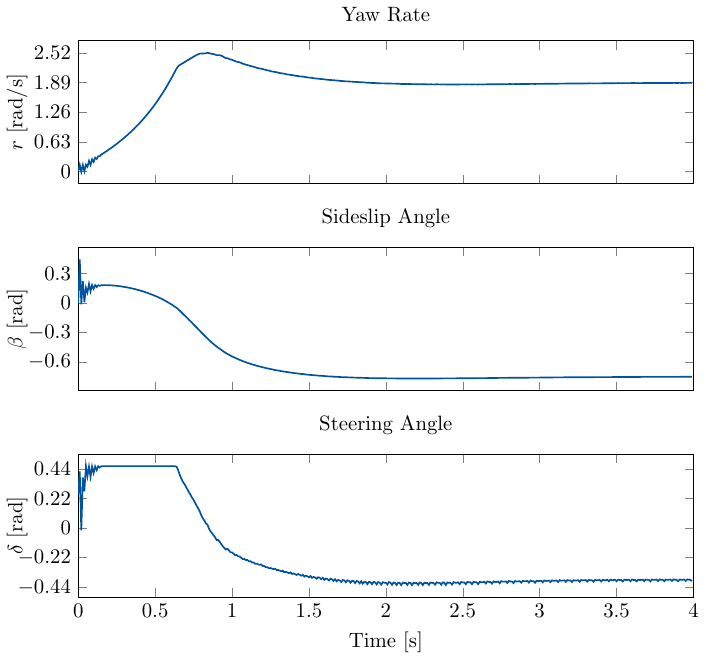}
    \caption*{(b)}
    \end{minipage}
    \vspace{-0.2cm}
    \caption{\yihan{Performance of the RWD drifting policy on a circular trajectory. (a) Paths for circular trajectory tracking. (b) Time histories of states ($r$, $\beta$) and the steering angle ($\delta$).}}
    \label{fig:rwd_comparison}
    \vspace{-0.4cm}
\end{figure}

\yihan{To validate the advantages of IWD's enhanced control authority demonstrated above, we also train policies on RWD configurations using identical hyperparameters. As shown in Fig.~\ref{fig:rwd_comparison}, the RWD system achieves stable circular drifting but requires approximately 1.5s to reach drift states compared to IWD's 1s, shows slightly larger deviations from the reference path, operates at smaller sideslip angles (-0.76 rad vs -0.85 rad), and demonstrates higher steering dependency.}

\yihan{Beyond circular trajectories, we evaluate the IWD system on more complex trajectory patterns.} The eight-shaped path adds complexity through drift direction reversals. As shown in Fig. \ref{fig:results_comparison}(a,c), the vehicle effectively tracks the reference path while maintaining mean sideslip angles of 47° and vehicle speeds of 1.81 m/s.
During the direction reversal phase (highlighted in red), the controller employs strategic differential wheel speeds: the front right wheel decelerates to 1.0 m/s while the left increases to 5.2 m/s. The rear wheels show even larger differentials, with the right at 1.0 m/s and left reaching 7.0 m/s, creating the substantial yaw moment needed for rotation.
The transition can be characterized by three phases: right wheel deceleration (3.2--3.4\,s), maximum wheel speed differential (3.4--3.6\,s), and convergence to a new drift equilibrium (3.7--3.9\,s), revealing the controller's ability to manage drift direction changes.

The variable-curvature track combines tight circles ($\kappa = 1.0$), gentle curves ($\kappa = 0.5$), and smooth transitions, inspired by \cite{goh2018controller,domberg_deep_2022}. The controller maintains a consistent sideslip angle around -50° as shown in Fig. \ref{fig:results_comparison}(c), despite the changing path geometry. For the red segment in Fig. \ref{fig:results_comparison}(b), wheel speed differentials adapt smoothly to local curvature variations, demonstrating stable drift behavior throughout the trajectory.

The simulation results demonstrate effective autonomous drifting across multiple trajectory complexities, from basic circular tracking to challenging maneuvers requiring direction reversals. \yihan{The comparative analysis demonstrates that IWD systems achieve better tracking precision and faster drift initiation through additional yaw moments generated by differential wheel speed control.}

\subsection{Real-World Validation}

\begin{figure}[!b]
    \centering
    \includegraphics[width=0.8\columnwidth]{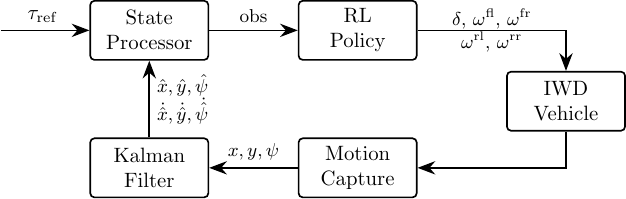}
    \caption{Real-world deployment architecture.}
    \label{fig:deploy}
\end{figure}

We implement our approach on a 1/10 scale IWD RC car platform mentioned in Section~\ref{sec:hardware}. The trained policy transfers directly to the real platform without any additional fine-tuning, executing at 100 Hz on the onboard computer. The deployment architecture is illustrated in Fig. \ref{fig:deploy}. Video demonstrations of the real-world experiments are available at \url{https://youtu.be/6ovkdj5\_1Yk}.

\begin{figure*}[t]
    \centering
    \begin{minipage}[c]{0.48\textwidth}
        \centering
        \text{Simulation}\\
        \vspace{0.2cm}
        \begin{minipage}[c]{0.48\linewidth}
            \centering
            \includegraphics[width=0.9\columnwidth]{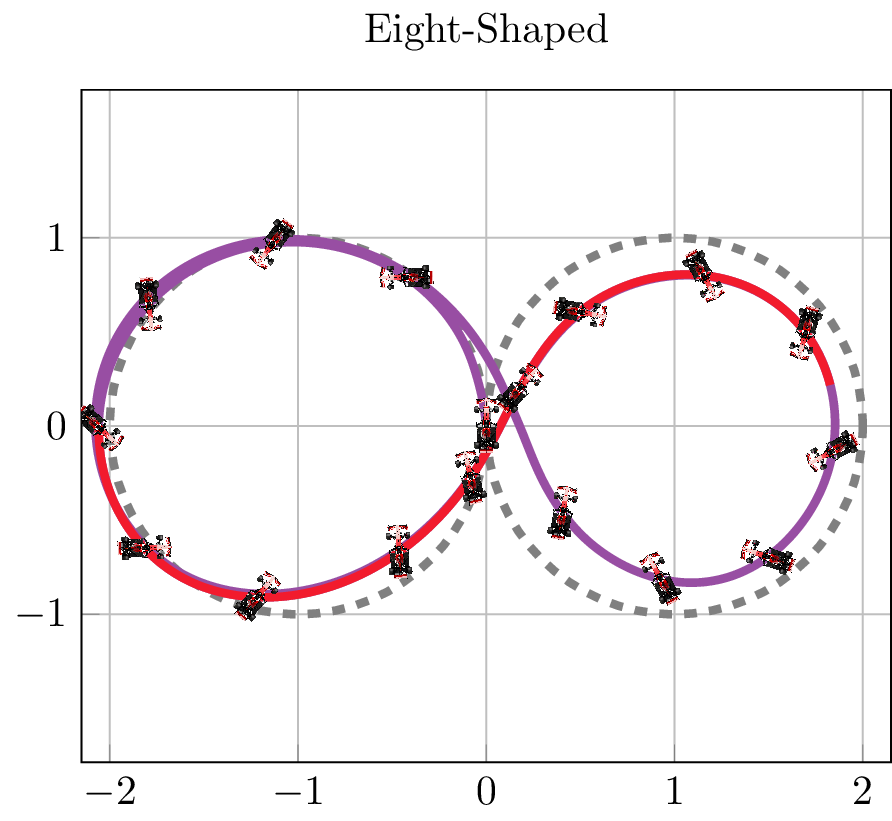}
            \vspace{0.2cm}
            \includegraphics[width=0.9\columnwidth]{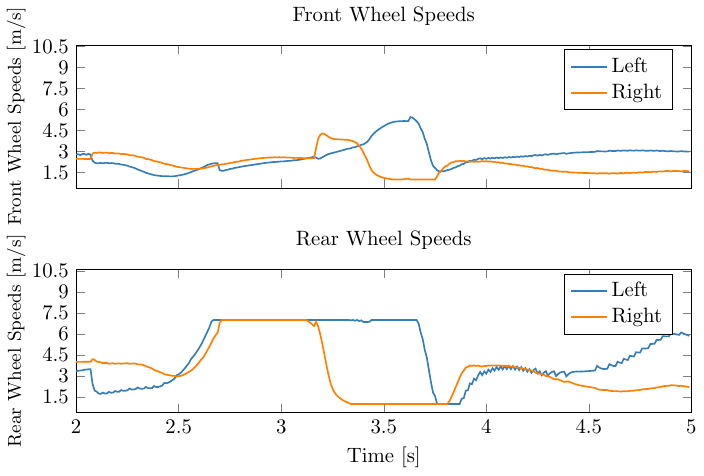}
            \vspace{-0.3cm}
            \caption*{(a)}
        \end{minipage}%
        \hfill
        \begin{minipage}[c]{0.48\linewidth}
            \centering
            \includegraphics[width=0.9\columnwidth]{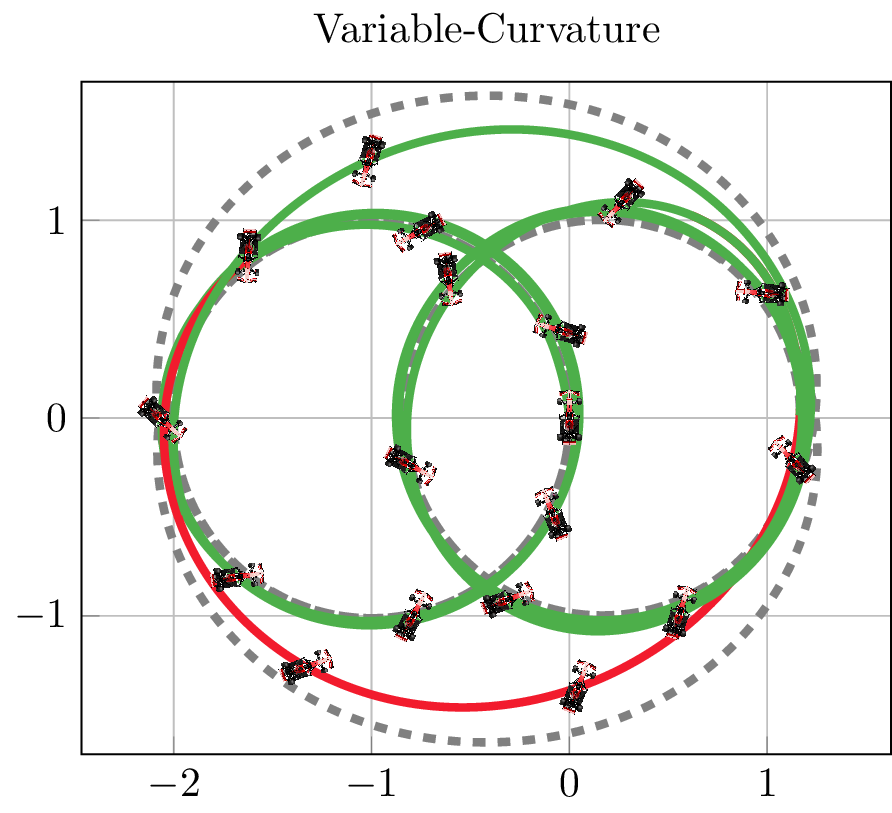}
            \vspace{0.2cm}
            \includegraphics[width=0.9\columnwidth]{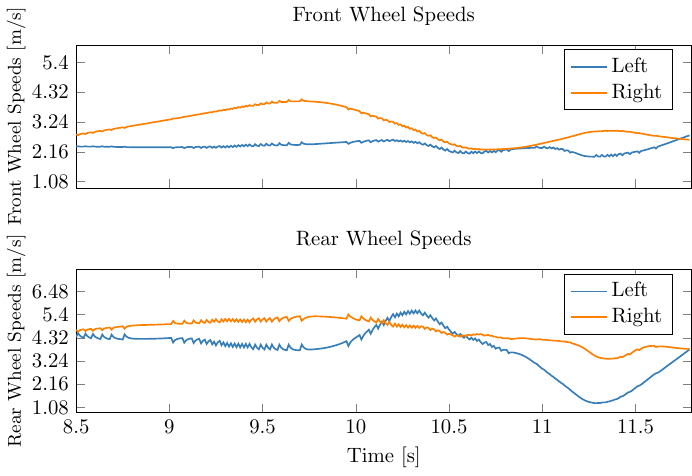}
            \vspace{-0.3cm}
            \caption*{(b)}
        \end{minipage}
        \includegraphics[width=\linewidth]{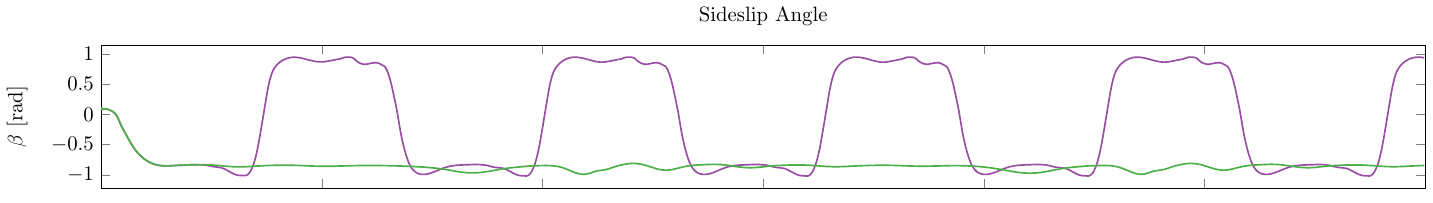}
        \includegraphics[width=\linewidth]{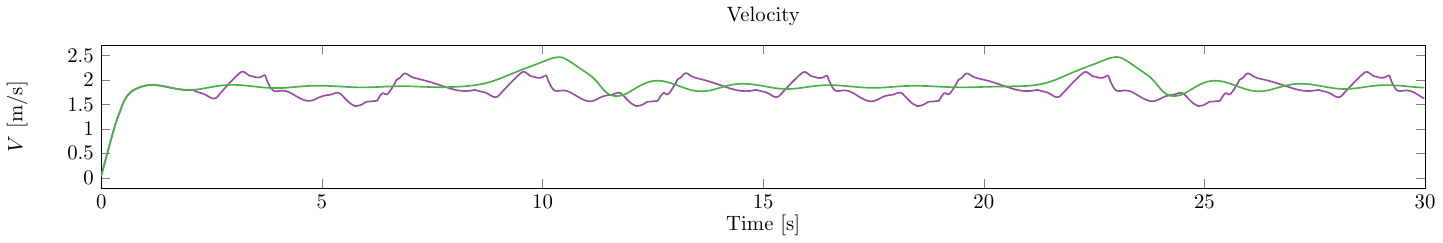}
        \vspace{-0.5cm}
        \caption*{(c)}
    \end{minipage}%
    \hfill
    \begin{minipage}[c]{0.48\textwidth}
        \centering
        \text{Real-World}\\
        \vspace{0.2cm}
        \begin{minipage}[c]{0.48\linewidth}
            \centering
            \includegraphics[width=0.9\columnwidth]{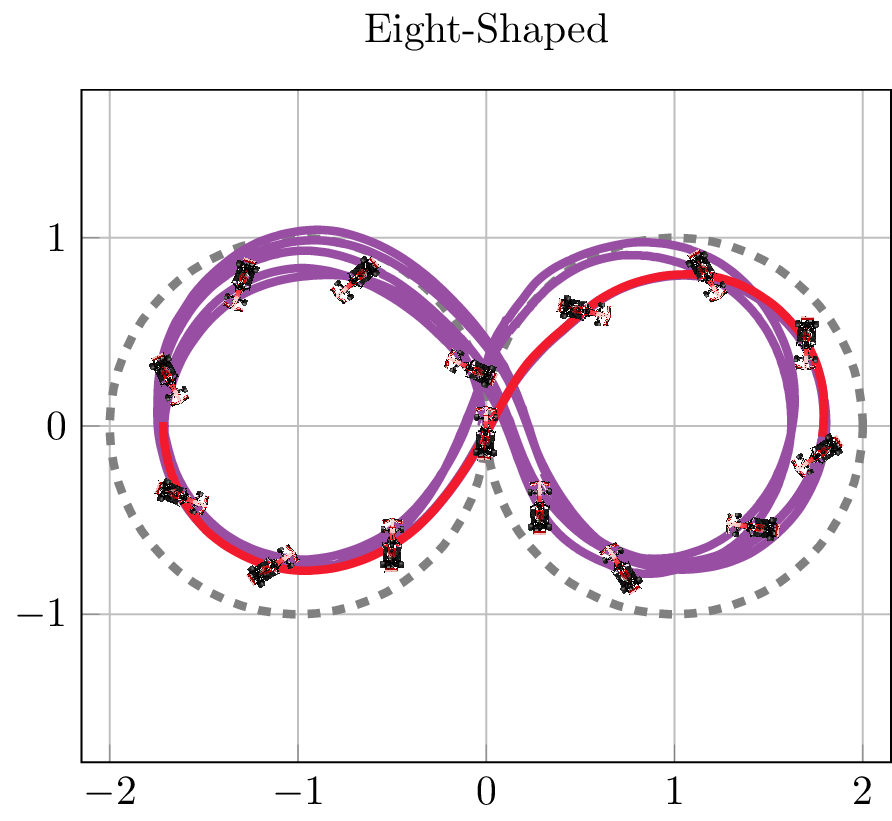}
            \vspace{0.2cm}
            \includegraphics[width=0.9\columnwidth]{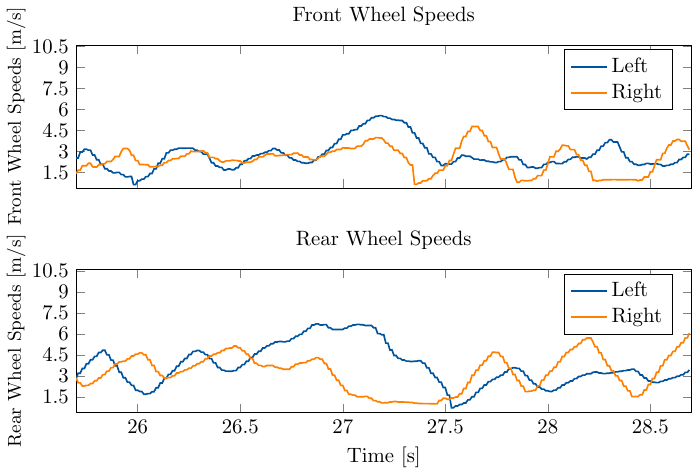}
            \vspace{-0.3cm}
            \caption*{(d)}
        \end{minipage}%
        \hfill
        \begin{minipage}[c]{0.48\linewidth}
            \centering
            \includegraphics[width=0.9\columnwidth]{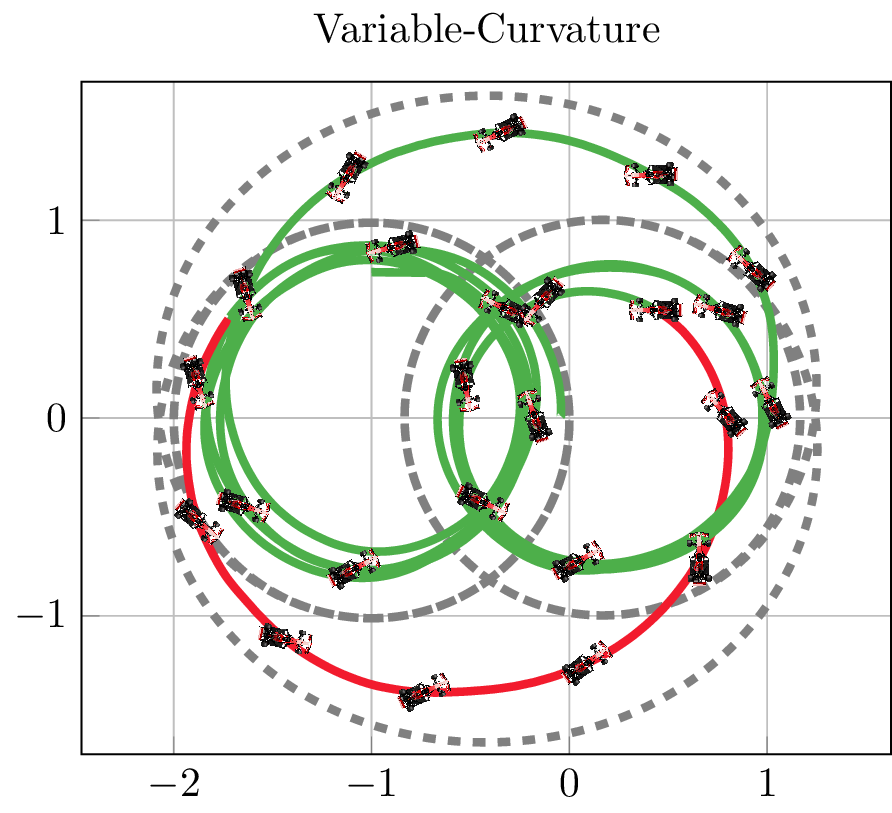}
            \vspace{0.2cm}
            \includegraphics[width=0.9\columnwidth]{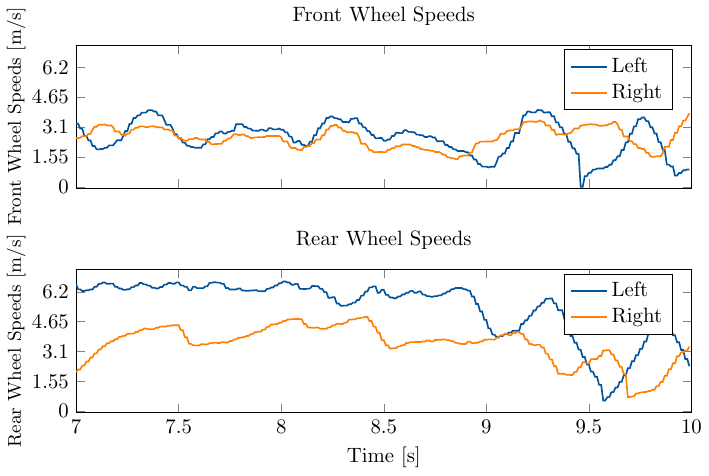}
            \vspace{-0.3cm}
            \caption*{(e)}
        \end{minipage}
        \includegraphics[width=\linewidth]{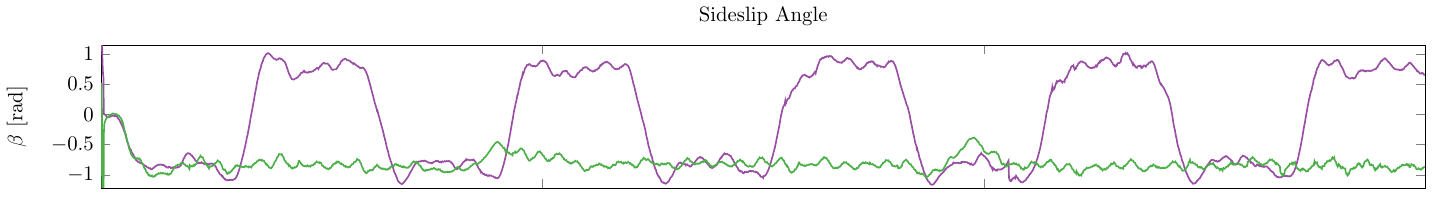}
        \includegraphics[width=\linewidth]{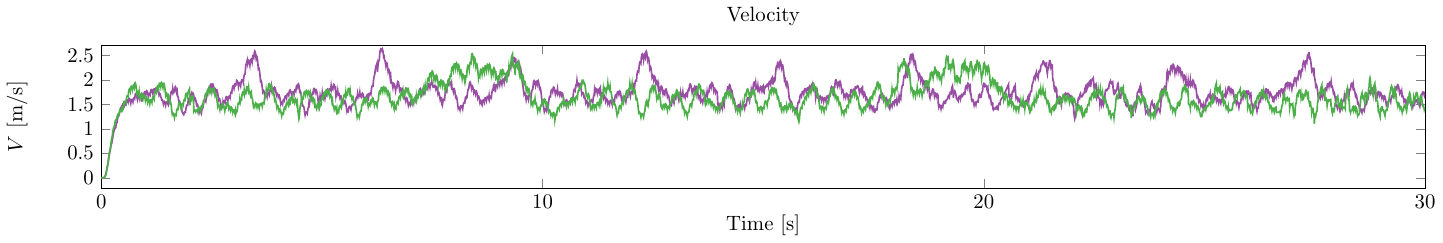}
        \vspace{-0.5cm}
        \caption*{(f)}
    \end{minipage}
    \caption{IWD Performance comparison between simulation (left) and real-world experiments (right). Path tracking results for (a,d) eight-shaped path and (b,e) variable-curvature track, showing actual vehicle paths (purple/green) against reference paths (gray) with corresponding wheel speed time histories for highlighted segments (in red). (c,f) Comparison of sideslip angles and velocities for both trajectories (purple for eight-shaped path, green for variable-curvature track).}
    \label{fig:results_comparison}
    \vspace{-0.5cm}
\end{figure*}

To validate our approach, we focus on the eight-shaped path and the variable-curvature track, as shown in Fig.~\ref{fig:results_comparison}.
Real-world experiments demonstrate remarkable consistency with simulation in maintaining stable sideslip angles and velocities, though with slightly more conservative trajectories characterized by smaller turning radius. For the eight-shaped path, during direction reversals as highlighted in Fig.~\ref{fig:results_comparison}(d), the controller achieves rapid drift transitions by dramatically increasing left wheel speeds, effectively exploiting the IWD configuration for yaw control. On the variable-curvature track, while in simulation drifting is maintained through higher right wheel speeds relative to left, the physical system achieves stable drifting through an opposite strategy of higher left wheel speeds, yet successfully maintains consistent drift states throughout the trajectory with the compromise of smaller sideslip angles. To quantify the tracking performance, we compute the root mean square error (RMSE) of position tracking for both simulation and real-world experiments, as summarized in Table~\ref{tab:rmse_comparison}.

\begin{table}[!b]
    \renewcommand{\arraystretch}{1.2}
    \caption{Position Error RMSE (mean $\pm$ std, n=6)}
    \label{tab:rmse_comparison}
    \centering
    \begin{tabular}{lcc}
    \toprule
    \textbf{Environment} & \textbf{Eight-Shaped Path} & \textbf{Variable-Curvature Path} \\
    \midrule
    Simulation & \yihan{$0.138 \pm 0.001$} m & \yihan{$0.075 \pm 0.002$} m \\
    Real-World & \yihan{$0.221 \pm 0.022$} m & \yihan{$0.231 \pm 0.037$} m \\
    \bottomrule
    \end{tabular}
\end{table}

To further evaluate our approach, we test the vehicle on a path inspired by the Olympic rings where sustained drifting must be maintained through five tangent circles while repeatedly reversing drift direction at tangent points. The consistent state transitions shown in Fig.~\ref{fig:top-down} demonstrate the controller's ability to maintain stable drift through multiple reversals, highlighting the reliability of our approach.
\section{Conclusion}
\label{sec:conclusion}
\yihan{This paper presents a reinforcement learning approach for autonomous drifting control using Individual Wheel Drive vehicles. The proposed framework achieves effective sim-to-real transfer through GPU-accelerated parallel simulation and systematic domain randomization, enabling successful deployment without fine-tuning.

Experimental validation demonstrates effectiveness across diverse drifting scenarios, including direction reversals and variable-curvature paths, with real-world experiments confirming consistent performance. IWD systems generate additional yaw moments through differential wheel speeds, enabling fast drift initiation and precise trajectory tracking.
The open-source platform and codebase facilitate further research in IWD vehicle control and autonomous drifting applications.
}

\revise{Future work could incorporate lateral load transfer and suspension dynamics to better capture weight redistribution effects during aggressive cornering, potentially enhancing policy robustness and performance in extreme drift scenarios.}

\appendix
\section{Ablation Study}
\revise{
To validate the effectiveness of our domain randomization strategy, we conduct ablation experiments by training policies with different randomization configurations. Each policy is evaluated on 100 trials of eight-shaped trajectories under challenging test conditions with broader parameter distributions: Pacejka parameters $B \in [0.2,3]$, $C \in [1.5,3]$, $D \in [0.2,0.5]$ and increased disturbance scaling.

\begin{table}[!t]
\renewcommand{\arraystretch}{1.2}
\caption{Domain Randomization Ablation Study Results}
\label{tab:ablation_results}
\centering
\footnotesize
\begin{tabular}{lccc}
\toprule
Method & \begin{tabular}{c}Success\\Rate\end{tabular} & RMSE (m) & \begin{tabular}{c}Sideslip\\Angle (°)\end{tabular} \\
\midrule
Full randomization & 75.0\% & 0.146±0.053 & 0.81±0.09 \\
w/o tire rand. & 73.0\% & 0.158±0.054 & 0.88±0.09 \\
w/o init. state rand. & 46.0\% & 0.233±0.049 & 0.79±0.06 \\
w/o dynamic disturb. & 80.0\% & 0.159±0.062 & 0.76±0.10 \\
w/o traj. rand. & 73.0\% & 0.222±0.053 & 0.71±0.07 \\
\bottomrule
\end{tabular}
\end{table}
The results in Table~\ref{tab:ablation_results} demonstrate that initial state randomization is the most critical component, with its removal causing a 29\% drop in success rate and 60\% increase in tracking error. Trajectory randomization also significantly impacts performance, showing 52\% increase in RMSE when removed.

While removing dynamic disturbance shows slight simulation improvement (80\% vs 75\%), only the full randomization configuration successfully transfers to real-world deployment. This indicates that dynamic disturbance is essential for handling unmodeled dynamics in physical systems.
}

\bibliographystyle{IEEEtran}
\bibliography{ref.bib}

\begin{thebibliography}{10}
\providecommand{\url}[1]{#1}
\csname url@samestyle\endcsname
\providecommand{\newblock}{\relax}
\providecommand{\bibinfo}[2]{#2}
\providecommand{\BIBentrySTDinterwordspacing}{\spaceskip=0pt\relax}
\providecommand{\BIBentryALTinterwordstretchfactor}{4}
\providecommand{\BIBentryALTinterwordspacing}{\spaceskip=\fontdimen2\font plus
\BIBentryALTinterwordstretchfactor\fontdimen3\font minus
  \fontdimen4\font\relax}
\providecommand{\BIBforeignlanguage}[2]{{%
\expandafter\ifx\csname l@#1\endcsname\relax
\typeout{** WARNING: IEEEtran.bst: No hyphenation pattern has been}%
\typeout{** loaded for the language `#1'. Using the pattern for}%
\typeout{** the default language instead.}%
\else
\language=\csname l@#1\endcsname
\fi
#2}}
\providecommand{\BIBdecl}{\relax}
\BIBdecl

\bibitem{cai2020high}
P.~Cai, X.~Mei, L.~Tai, Y.~Sun, and M.~Liu, ``High-speed autonomous drifting
  with deep reinforcement learning,'' \emph{IEEE Robotics and Automation
  Letters}, vol.~5, no.~2, pp. 1247--1254, 2020.

\bibitem{betz2022autonomous}
J.~Betz, H.~Zheng, A.~Liniger, U.~Rosolia, P.~Karle, M.~Behl, V.~Krovi, and
  R.~Mangharam, ``Autonomous vehicles on the edge: A survey on autonomous
  vehicle racing,'' \emph{IEEE Open Journal of Intelligent Transportation
  Systems}, vol.~3, pp. 458--488, 2022.

\bibitem{goh2020toward}
J.~Y. Goh, T.~Goel, and J.~Christian~Gerdes, ``Toward automated vehicle control
  beyond the stability limits: drifting along a general path,'' \emph{Journal
  of Dynamic Systems, Measurement, and Control}, vol. 142, no.~2, p. 021004,
  2020.

\bibitem{yang2022hierarchical}
B.~Yang, Y.~Lu, X.~Yang, and Y.~Mo, ``A hierarchical control framework for
  drift maneuvering of autonomous vehicles,'' in \emph{2022 International
  Conference on Robotics and Automation (ICRA)}.\hskip 1em plus 0.5em minus
  0.4em\relax IEEE, 2022, pp. 1387--1393.

\bibitem{lu2023consecutive}
Y.~Lu, B.~Yang, J.~Li, Y.~Zhou, H.~Chen, and Y.~Mo, ``Consecutive inertia drift
  of autonomous rc car via primitive-based planning and data-driven control,''
  \emph{arXiv preprint arXiv:2306.12604}, 2023.

\bibitem{voser2010analysis}
C.~Voser, R.~Y. Hindiyeh, and J.~C. Gerdes, ``Analysis and control of high
  sideslip manoeuvres,'' \emph{Vehicle System Dynamics}, vol.~48, no.~S1, pp.
  317--336, 2010.

\bibitem{hindiyeh2014controller}
R.~Y. Hindiyeh and J.~Christian~Gerdes, ``A controller framework for autonomous
  drifting: Design, stability, and experimental validation,'' \emph{Journal of
  Dynamic Systems, Measurement, and Control}, vol. 136, no.~5, p. 051015, 2014.

\bibitem{weber2023modeling}
T.~P. Weber and J.~C. Gerdes, ``Modeling and control for dynamic drifting
  trajectories,'' \emph{IEEE Transactions on Intelligent Vehicles}, 2023.

\bibitem{djeumou2024one}
F.~Djeumou, T.~J. Lew, N.~Ding, M.~Thompson, M.~Suminaka, M.~Greiff, and
  J.~Subosits, ``One model to drift them all: Physics-informed conditional
  diffusion model for driving at the limits,'' in \emph{8th Annual Conference
  on Robot Learning}, 2024.

\bibitem{cutler_autonomous_2016}
M.~Cutler and J.~P. How, ``\BIBforeignlanguage{en}{Autonomous drifting using
  simulation-aided reinforcement learning},'' in
  \emph{\BIBforeignlanguage{en}{2016 {IEEE} {International} {Conference} on
  {Robotics} and {Automation} ({ICRA})}}.\hskip 1em plus 0.5em minus
  0.4em\relax IEEE, May 2016, pp. 5442--5448.

\bibitem{domberg_deep_2022}
F.~Domberg, C.~C. Wembers, H.~Patel, and G.~Schildbach,
  ``\BIBforeignlanguage{en}{Deep {Drifting}: {Autonomous} {Drifting} of
  {Arbitrary} {Trajectories} using {Deep} {Reinforcement} {Learning}},'' in
  \emph{\BIBforeignlanguage{en}{2022 {International} {Conference} on {Robotics}
  and {Automation} ({ICRA})}}.\hskip 1em plus 0.5em minus 0.4em\relax IEEE, May
  2022, pp. 7753--7759.

\bibitem{goh2016simultaneous}
J.~Y. Goh and J.~C. Gerdes, ``Simultaneous stabilization and tracking of basic
  automobile drifting trajectories,'' in \emph{2016 IEEE Intelligent Vehicles
  Symposium (IV)}.\hskip 1em plus 0.5em minus 0.4em\relax IEEE, 2016, pp.
  597--602.

\bibitem{meijer2024nonlinear}
S.~Meijer, A.~Bertipaglia, and B.~Shyrokau, ``A nonlinear model predictive
  control for automated drifting with a standard passenger vehicle,''
  \emph{arXiv preprint arXiv:2405.10859}, 2024.

\bibitem{10219145}
X.~Tian, S.~Yang, Y.~Yang, W.~Song, and M.~Fu, ``A multi-layer drifting
  controller for all-wheel drive vehicles beyond driving limits,''
  \emph{IEEE/ASME Transactions on Mechatronics}, vol.~29, no.~2, pp.
  1229--1239, 2024.

\bibitem{goh2018controller}
J.~Y. Goh, T.~Goel, and J.~C. Gerdes, ``A controller for automated drifting
  along complex trajectories,'' in \emph{14th International Symposium on
  Advanced Vehicle Control (AVEC 2018)}, vol.~7, 2018, pp. 1--6.

\bibitem{hu2024novel}
C.~Hu, L.~Xie, Z.~Zhang, and H.~Xiong, ``A novel model predictive controller
  for the drifting vehicle to track a circular trajectory,'' \emph{Vehicle
  System Dynamics}, pp. 1--30, 2024.

\bibitem{shi2023nonlinear}
Z.~Shi, H.~Chen, S.~Yu, R.~Findeisen, and H.~Guo, ``Nonlinear model predictive
  control for autonomous vehicle drifting,'' \emph{International Journal of
  Robust and Nonlinear Control}, 2023.

\bibitem{velenis2009steady}
E.~Velenis, E.~Frazzoli, and P.~Tsiotras, ``On steady-state cornering
  equilibria for wheeled vehicles with drift,'' in \emph{Proceedings of the 48h
  IEEE Conference on Decision and Control (CDC) held jointly with 2009 28th
  Chinese Control Conference}.\hskip 1em plus 0.5em minus 0.4em\relax IEEE,
  2009, pp. 3545--3550.

\bibitem{velenis2011steady}
E.~Velenis, D.~Katzourakis, E.~Frazzoli, P.~Tsiotras, and R.~Happee,
  ``Steady-state drifting stabilization of rwd vehicles,'' \emph{Control
  Engineering Practice}, vol.~19, no.~11, pp. 1363--1376, 2011.

\bibitem{joa2020new}
E.~Joa, H.~Cha, Y.~Hyun, Y.~Koh, K.~Yi, and J.~Park, ``A new control approach
  for automated drifting in consideration of the driving characteristics of an
  expert human driver,'' \emph{Control Engineering Practice}, vol.~96, p.
  104293, 2020.

\bibitem{djeumou2023autonomous}
F.~Djeumou, J.~Goh, U.~Topcu, and A.~Balachandran, ``Autonomous drifting with 3
  minutes of data via learned tire models,'' in \emph{2023 International
  Conference on Robotics and Automation (ICRA)}, 2023.

\bibitem{deisenroth2011pilco}
M.~Deisenroth and C.~E. Rasmussen, ``Pilco: A model-based and data-efficient
  approach to policy search,'' in \emph{Proceedings of the 28th International
  Conference on machine learning (ICML-11)}, 2011, pp. 465--472.

\bibitem{bhattacharjee2018autonomous}
S.~Bhattacharjee, K.~D. Kabara, R.~Jain, and K.~Kabara, ``Autonomous drifting
  rc car with reinforcement learning,'' \emph{Dept. Comput. Sci., Univ. Hong
  Kong, Tech. Rep}, 2018.

\bibitem{gonzales2018planning}
J.~M. Gonzales, \emph{Planning and control of drift maneuvers with the Berkeley
  autonomous race car}.\hskip 1em plus 0.5em minus 0.4em\relax University of
  California, Berkeley, 2018.

\bibitem{karaman2017project}
S.~Karaman, A.~Anders, M.~Boulet, J.~Connor, K.~Gregson, W.~Guerra, O.~Guldner,
  M.~Mohamoud, B.~Plancher, R.~Shin \emph{et~al.}, ``Project-based,
  collaborative, algorithmic robotics for high school students: Programming
  self-driving race cars at mit,'' in \emph{2017 IEEE integrated STEM education
  conference (ISEC)}.\hskip 1em plus 0.5em minus 0.4em\relax IEEE, 2017, pp.
  195--203.

\bibitem{srinivasa2019}
S.~S. Srinivasa, P.~Lancaster, J.~Michalove, M.~Schmittle, C.~Summers,
  M.~Rockett, J.~R. Smith, S.~Choudhury, C.~Mavrogiannis, and F.~Sadeghi,
  ``Mushr: A low-cost, open-source robotic racecar for education and
  research,'' 2019.

\bibitem{Hart2014}
K.~Hart, C.~Montella, G.~Petitpas, D.~Schweisinger, A.~Shariati, B.~Sourbeer,
  T.~Trephan, and J.~Spletzer, ``{RoSCAR},'' in \emph{Proceedings of the 2014
  workshop on Mobile augmented reality and robotic technology-based systems -
  {MARS} {\textquotesingle}14}.\hskip 1em plus 0.5em minus 0.4em\relax {ACM}
  Press, 2014.

\bibitem{okelly2019}
M.~O'Kelly, V.~Sukhil, H.~Abbas, J.~Harkins, C.~Kao, Y.~V. Pant, R.~Mangharam,
  D.~Agarwal, M.~Behl, P.~Burgio, and M.~Bertogna, ``F1/10: An open-source
  autonomous cyber-physical platform,'' 2019.

\bibitem{milani2021vehicle}
S.~Milani, H.~Marzbani, and R.~N. Jazar, ``Vehicle drifting dynamics: discovery
  of new equilibria,'' \emph{Vehicle System Dynamics}, pp. 1--26, 2021.

\bibitem{makoviychuk2021isaac}
V.~Makoviychuk, L.~Wawrzyniak, Y.~Guo, M.~Lu, K.~Storey, M.~Macklin,
  D.~Hoeller, N.~Rudin, A.~Allshire, A.~Handa \emph{et~al.}, ``Isaac gym: High
  performance gpu-based physics simulation for robot learning,'' \emph{arXiv
  preprint arXiv:2108.10470}, 2021.

\bibitem{o2020f1tenth}
M.~O'Kelly, H.~Zheng, D.~Karthik, and R.~Mangharam, ``F1tenth: An open-source
  evaluation environment for continuous control and reinforcement learning,''
  \emph{Proceedings of Machine Learning Research}, vol. 123, 2020.

\bibitem{paszke2017automatic}
A.~Paszke, S.~Gross, S.~Chintala, G.~Chanan, E.~Yang, Z.~DeVito, Z.~Lin,
  A.~Desmaison, L.~Antiga, and A.~Lerer, ``Automatic differentiation in
  pytorch,'' 2017.

\bibitem{schulman2017proximal}
J.~Schulman, F.~Wolski, P.~Dhariwal, A.~Radford, and O.~Klimov, ``Proximal
  policy optimization algorithms,'' \emph{arXiv preprint arXiv:1707.06347},
  2017.

\bibitem{schulman2015trust}
J.~Schulman, S.~Levine, P.~Abbeel, M.~Jordan, and P.~Moritz, ``Trust region
  policy optimization,'' in \emph{International conference on machine
  learning}.\hskip 1em plus 0.5em minus 0.4em\relax PMLR, 2015, pp. 1889--1897.

\end{thebibliography}



\end{document}